\documentclass[letterpaper, 10 pt, conference]{ieeeconf}
\IEEEoverridecommandlockouts
\usepackage{cite}
\usepackage{amsmath,amssymb,amsfonts}
\usepackage[linesnumbered, ruled]{algorithm2e}

\usepackage{algorithmic}
\SetKwRepeat{Do}{do}{while}%
\usepackage{graphicx}
\usepackage{textcomp}
\usepackage{xcolor}
\usepackage{float}
\usepackage{subfigure}
\usepackage{subcaption}
\usepackage{mathrsfs}
\usepackage{multicol}
\usepackage{multirow}
\usepackage{float}
\usepackage{booktabs}
\usepackage{caption}
\usepackage{setspace}

\title{\LARGE \bf
A Learning-based Control Methodology for Transitioning VTOL UAVs
}

\author{Zexin Lin$^{1, 2, *}$, Yebin Zhong$^{1, 2, *}$, Hanwen Wan$^{1, 2}$, Jiu Cheng$^{1, 2}$, Zhenglong Sun$^{1, 2}$,\\ and Xiaoqiang Ji$^{1, 2, \dag}$, \emph{IEEE Member}
\thanks{$^{1}$The School of Science and Engineering, The Chinese University of Hong Kong, Shenzhen, Shenzhen, 2001 Longxiang Boulevard, China.}
\thanks{$^{2}$The Shenzhen Institute of Artificial Intelligence and Robotics for Society Shenzhen, China.}
\thanks{$^{*}$These authors contributed to the work equally and should be regarded as co-first authors.} 
\thanks{$^{\dag}$Corresponding author is Xiaoqiang Ji whose e-mail is {\tt\small jixiaoqiang@cuhk.edu.cn}.}
}

\begin{document}
\maketitle
\thispagestyle{empty}
\pagestyle{empty}
\begin{abstract}
Transition control poses a critical challenge in Vertical Take-Off and Landing Unmanned Aerial Vehicle (VTOL UAV) development due to the tilting rotor mechanism, which shifts the center of gravity and thrust direction during transitions. Current control methods' decoupled control of altitude and position leads to significant vibration, and limits interaction consideration and adaptability. In this study, we propose a novel coupled transition control methodology based on reinforcement learning (RL) driven controller. Besides, contrasting to the conventional phase-transition approach, the ST3M method demonstrates a new perspective by treating cruise mode as a special case of hover. We validate the feasibility of applying our method in simulation and real-world environments, demonstrating efficient controller development and migration while accurately controlling UAV position and attitude, exhibiting outstanding trajectory tracking and reduced vibrations during the transition process.

\end{abstract}


\section{Introduction}
UAVs, which serve a multitude of roles from surveillance to disaster relief, have become integral to modern operations. Among the diverse UAV types, VTOL UAVs stand out for their hybrid capabilities, combining the best of fixed-wing and rotary-wing attributes, enabling vertical take-offs and landings, hovering, and cruising \cite{bulka2018}. This versatility allows VTOL UAVs to operate without the need for runways, and offer an enhanced balance of energy efficiency, maneuverability, and safety \cite{liang2023tailsitter}. The potential for transporting industry and related enterprises based on VTOL UAVs is surging.

\begin{figure}[htbp]
    \centering
    \includegraphics[width=\linewidth]{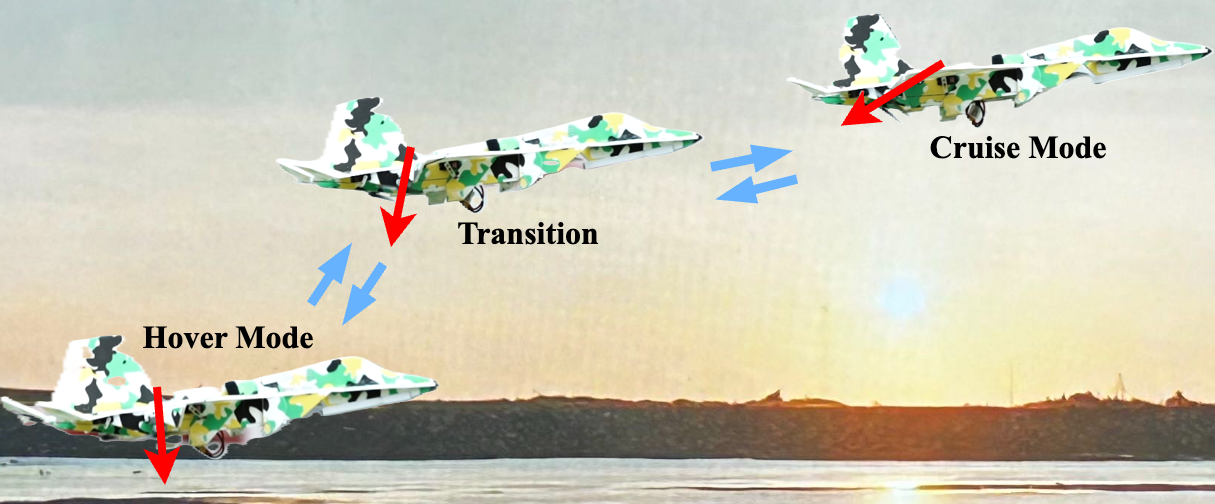}
    \caption{Tri-rotor VTOL UAV transition.}
    \label{fig:intro_fig}
\end{figure}

In this paper, our main interest is in the transition maneuvering problem of VTOL UAVs, which is a complex regime between hover and cruise flight modes as shown in Fig. \ref{fig:intro_fig}. Operating the tilting rotor mechanism affects system stability and is challenging as thrust direction and gravity center shift continuously \cite{lyu2017}. Meticulously engineered VTOL UAVs and control algorithms are essential to maintain stable posture and prevent tilting-induced capsizing \cite{zhou2017}. Stability is crucial for practical applications, especially in sectors like passenger and delicate cargo transport that require smooth handling \cite{wang2017}.

Recent years have seen marked progress on both the vehicle design and control fronts that are directly relevant to transition flight. On the airframe side, novel tilt-rotor and wings/tilt-wing configurations have been developed and validated experimentally or via high-fidelity simulation. For instance, a tilt-wing VTOL UAV design demonstrated successful transition flight dynamics and control simulation from hover to cruise and back \cite{DaudFilho2023_tiltwing_transition}. Another example is a new tilt-rotor UAV with a “swashplateless-elevon” actuation scheme that simplifies mechanical structure yet still achieves stable attitude control through transition by motor speed modulation instead of conventional control surfaces \cite{Mimouni2023_tiltrotor_no_surfaces}.

On the control-methodology side, researchers have moved beyond classical PD/PID and purely model-based MPC approaches. A recent study applied a backstepping sliding-mode thrust-vectoring controller to a novel tilt-rotor UAV, enhancing robustness during transition and improving disturbance rejection \cite{Yu2023_thrust_vectoring_tiltrotor}. Another line of work uses unified MPC strategies that handle both hover and forward-flight modes without mode-switching, fully exploiting actuator redundancy to ensure smooth mode transition and graceful failure handling \cite{Chen2024_unifiedMPC_tiltrotor}. Complementing model-based methods, learning-based controllers have also matured: a neural-network controller trained via imitation learning to replicate MPC behavior was shown capable of controlling the longitudinal axis of a tilt-rotor UAV across multiple transitions efficiently \cite{Ducard2024_NNC_tiltrotor}.

Additionally, improved aerodynamic modeling and transition-phase aerodynamic analyses provide deeper understanding of the flow regimes during conversion, informing safer and more efficient transition corridor design \cite{Wang2023_aero_sim_tiltrotor}. And as operations expand to more challenging environments (e.g. shipborne VTOL), high-precision transition trajectory optimization and control methods have been proposed recently to meet stringent safety and performance requirements \cite{Zhu2025_shipborne_VTOL_transition}.

Despite these advances, important gaps remain. Many model-based controllers still rely on decoupled altitude/position loops that can produce large attitude excursions during transitions \cite{yang2018,bauersfeld2021}; purely learning-based controllers often require extensive training data and careful sim-to-real handling; and integrated approaches that retain control-theoretic guarantees while leveraging data-driven components remain active research frontiers. Combined hardware–software co-design of tilting mechanisms, sensing suites, and control laws is increasingly recognized as necessary for reliable, safe, smooth transitions in real-world, disturbed, or cluttered environments.

The phase transition between different modes in VTOL UAVs requires precise coordination of sensors and actuators. Existing methods — such as proportional-derivative (PD) \cite{pd} control and proportional-integral-derivative (PID) \cite{PID1, PID2, PID3} control — have limitations due to their reliance on accurate mathematical models and sensitivity to uncertainties. Model predictive control (MPC) \cite{mpc1,mpc2,mpc3,mpc-vtol} offers sophisticated approaches but heavily depends on model accuracy, which is challenging in rapidly changing environments. Also, the Gauss pseudo-spectral method (GPM) \cite{gpm1,gpm2} faces similar limitations due to computational cost and model requirements. Model-based methods often decouple altitude and position control, leading to drastic changes in attitude angle during VTOL UAV phase transitions \cite{yang2018,bauersfeld2021}, preventing seamless switching.

RL controllers have the potential to generalize learning from one task to another, providing more flexibility in applying learned control policies to new scenarios \cite{openai_generalization}. A recent study demonstrated the potential of a mode-free, model-agnostic neural network controller for solving hybrid UAV transition flight problems \cite{rl5}. The main limitation of online learning is the slow learning speed of transition operation compared to hovering operation.

This paper proposed a novel learning-based VTOL UAV transition control methodology by treating the transition problem as a point-to-point hover point followed by coupled controls of altitude and position. It includes 2 stages: (i) the training method of the dynamic hovering controller and (ii) the optimization method of the target trajectory following. The contributions of this work are listed as follows.

\begin{itemize}
\item[$\bullet$] A new coupled VTOL UAV control methodology ST3M is proposed and implemented.

\item[$\bullet$] Smooth and stable transition control during stable point-to-point maneuvers.

\item[$\bullet$] Multi-stage progressive training paradigm for VTOL UAV RL controller to achieve smooth transition.

\end{itemize}

The rest of this paper is organized as follows. Section II introduces the system design including the hardware platform, simulation environment, and communication interface. Section III introduces the methodology. Section IV presents the simulation and experimental results analysis. Section V concludes the paper and discusses future work.

\section{System design}
This section provides the components of the physical and algorithm verification platform based on the previous framework \cite{ours}, which implements the experimental prototype and establishes a high-accuracy digital twin platform.

\begin{figure}
  \centering
    \includegraphics[width=.8\linewidth]{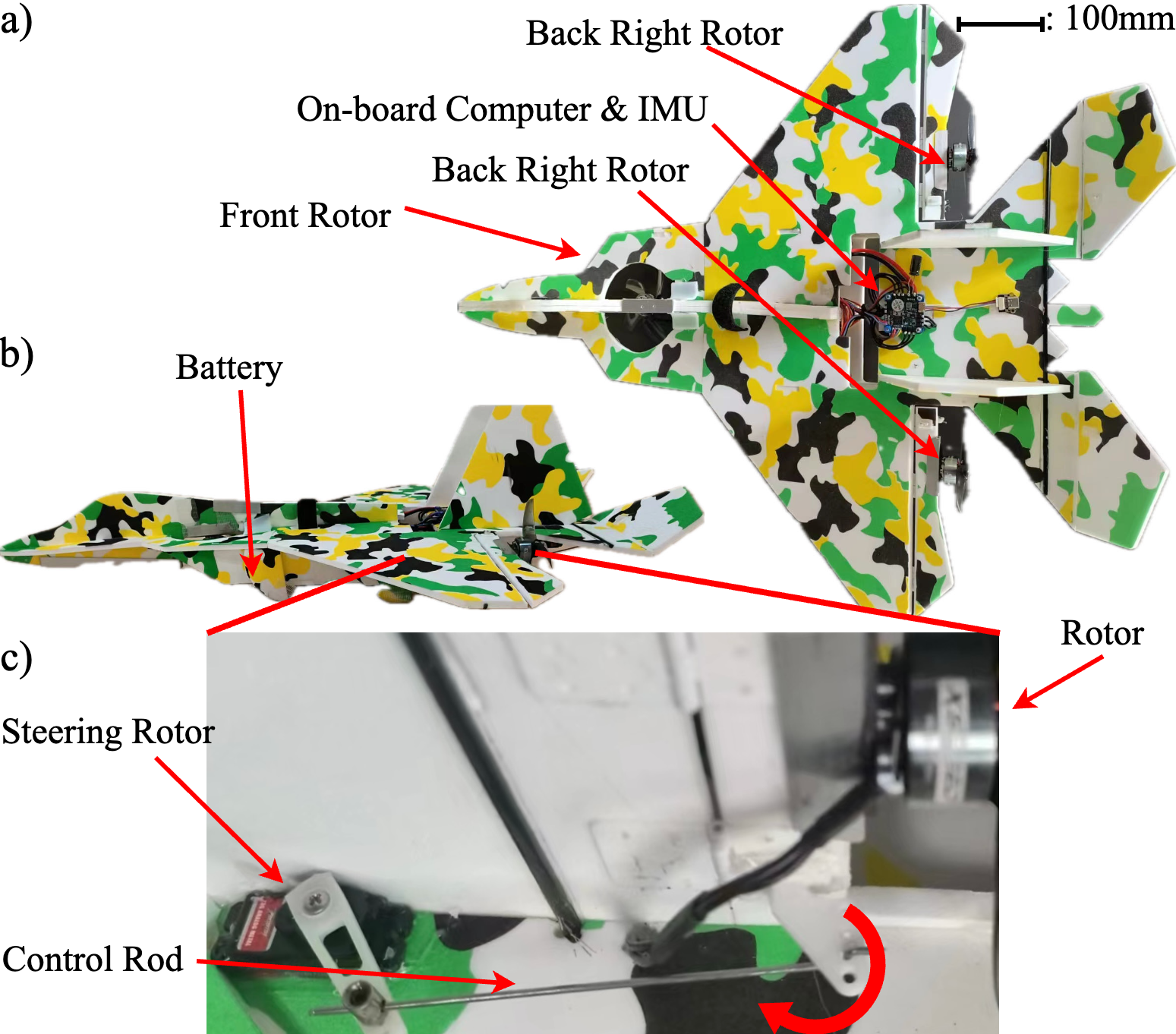}
    \caption{Tri-rotor VTOL UAV design.}
    \label{fig:hardware}
\end{figure}

\subsection{UAV Structure}
In the VTOL UAV, 3 SPARKHOBBY KV2750 motors were used to power the model. As shown in Fig. \ref{fig:hardware}b, 1 motor was used in the middle of the model to provide the main lift thrust. 2 servos and 2 FPV rotors were mounted on the back of two wings to provide dynamic thrust by pull or push the control rod and control the flaps as shown in Fig. \ref{fig:hardware}c, the tilting mechanism is constructed with an EMAX S08MA II servo, a control rod, and a flip-able flap with a thrust motor attached to it.

The onboard computer comprises a Teensy 4.0 board and an Aocoda-RC Electronic Speed Control (ESE). The Teensy board was used to control the servos and connected to the ESC. Three motors were connected to the ESC to control the rotating speed; thus, the thrusting power was controlled. A 4s FPV battery pack was put in the middle of the plane to have a better center of mass for the overall plane and easier to power the whole system. Detailed information about the hardware component specifications is listed in Table \ref{tab:specifications}.

\begin{table}[h]
    \centering
    \caption{Specifications of the VTOL UAV}
    \begin{tabular}{ccc}
         \toprule
          Item & Value\\
         \midrule
         Body Length & 700 mm\\
         Wing Span & 500 mm\\
         Total Weight (With Battery) & 1 kg\\
         Thrust Weight Ratio & 4\\
         Cruise Speed & 25 m/s\\
         Cruise Current & 3 A\\
         Hover Current & 5.8 A\\
         Battery & 4S, 1300 mAh\\
         \bottomrule
    \end{tabular}
    \label{tab:specifications}
\end{table}

\begin{figure*}[htbp]
\centering
\centerline{\includegraphics[width=\linewidth]{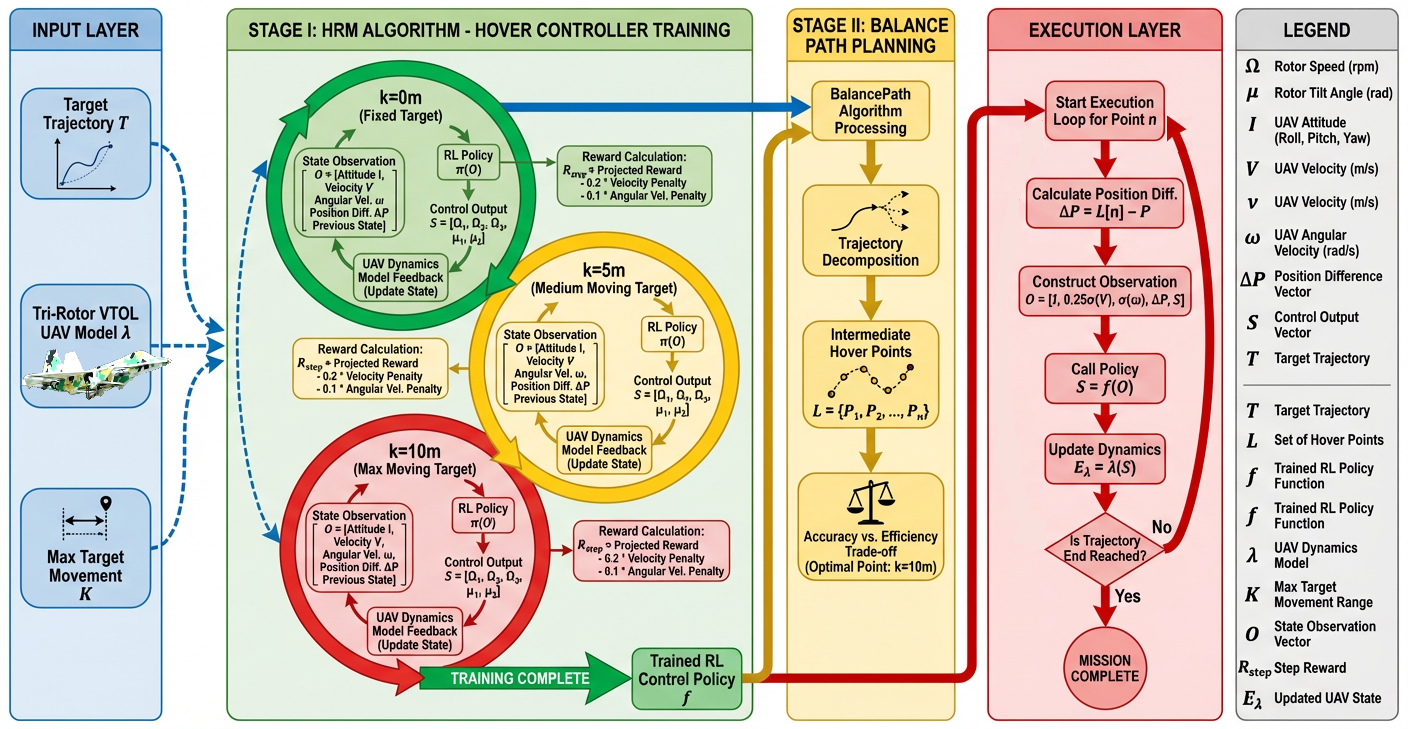}}\caption{The pipeline of the ST3M method.}
\label{fig:methodology}
\end{figure*}

\subsection{Simulation Environment}
\begin{figure}[htbp]
\centering
\centerline{\includegraphics[width=\linewidth]{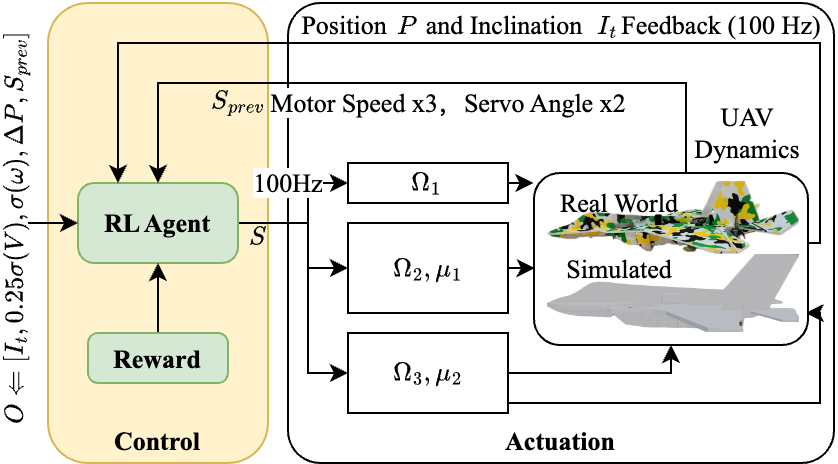}}\caption{RL control interface.}
\label{fig:rl_controller}
\end{figure}

The controlling RL agent was trained inside a simulation environment developed using Unity Engine and Unity ML-Agent Framework\cite{unity} to train and validate the performance of the trained agents before deploying them in the actual physical model. The simulated model has physical properties such as weight, size, and shape that are accurately replicated from the real-world hardware VTOL UAV, which we developed in the previous research\cite{ours}. As shown in Fig. \ref{fig:rl_controller}, the simulated model also shares identical interfaces with the real-world model.

\subsection{Communication Interface}
Deploying the RL controller onboard VTOL UAVs remains challenging due to computation power, cost, and power consumption limitations. Hence, we adopt a remote control approach. The neural network runs on a server, transmitting control commands via Bluetooth to the VTOL UAV. The UAV processes 5 floating-point numbers—representing rotor and servo control commands—received at a 100Hz frequency. Simultaneously, it sends acceleration data back to the server. Offloading the work to the server reduces computation power and increases battery life significantly, making real-world experiments feasible.
\section{Problem Statement}

\begin{figure}[htbp]
    \centering
    \includegraphics[width=.9\linewidth]{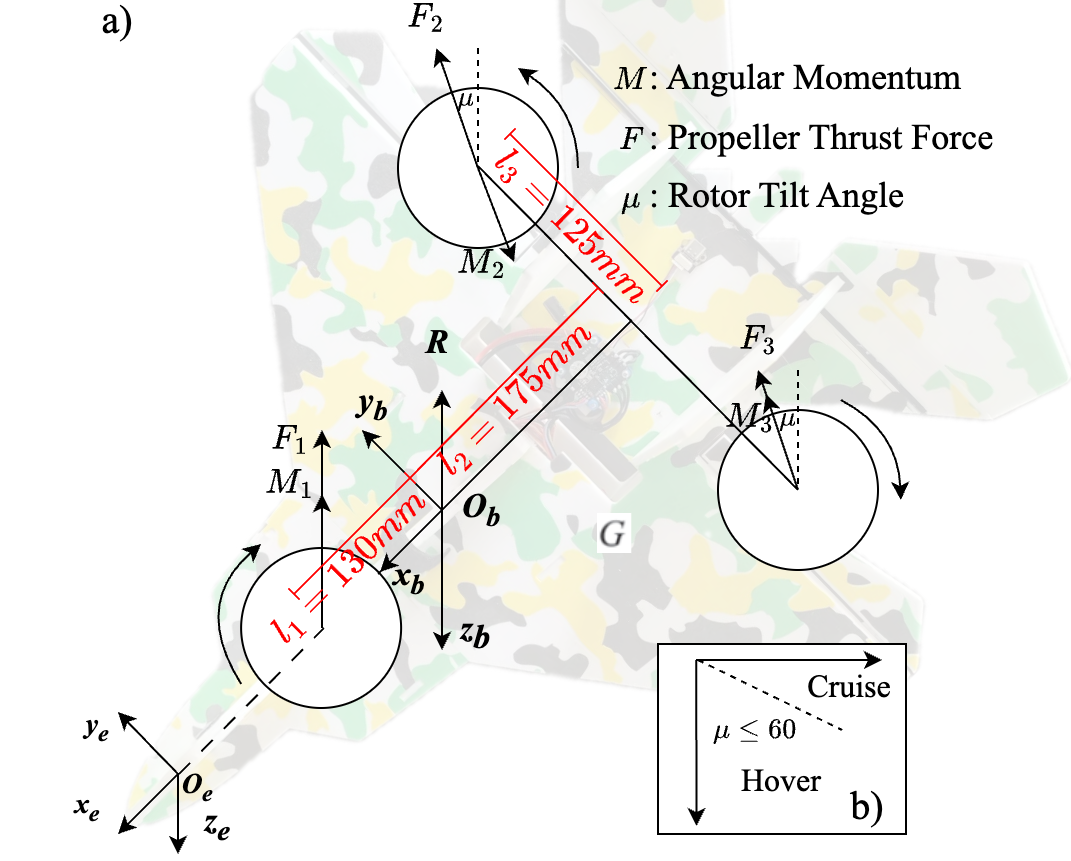}
    \caption{a) shows the tri-rotor VTOL UAV configuration in the earth (e) and body (b) axis systems; b) defines cruise and hover modes.}
    \label{fig:dynamic}
\end{figure}

\textbf{Notation.} The notation used in the following sections is fairly standard, i.e., $\textbf{1}_{m \times n}$ denotes a matrix whose elements are all 1, $Rand(\cdot)$ denotes the random operator.

\textbf{Assumptions.} The hover mode and cruise mode are defined as $\mu \leq 60^\circ$ and $\mu > 60^\circ$ as shown in Fig. \ref{fig:dynamic}b. The fan lift is ignored due to the low cruise speed.

Transition flight control can be considered as a reference trajectory tracking problem. To solve the stability during the transition flight, we consider the tracking problem from a reference trajectory to a series of reference hover points as:
\begin{align}
\begin{aligned}
&\min \ J=\sum e^i, i=1,2,...,N\\
& s.t. \ e^i = \left[ \begin{array}{c}
P^i-P^i_d \\
I^i-I^i_d \\
\end{array} \right]^\mathrm{T} \left[ \begin{matrix}
Q_1 & 0 \\
0 & Q_2 \\
\end{matrix} \right] \left[ \begin{array}{c}
P^i_i-P^i_d \\
I^i_i-I^i_d \\
\end{array} \right] \\
& 0 \le ||\dot P^{i+1}-\dot P^i||, ||\dot I^{i+1}-\dot I^i|| \leq \epsilon1_{max}, \epsilon2_{max},
\label{eq:problem}
\end{aligned}
\end{align}
where $P^i_d$, $I^i_d$, $Q_1$, and $Q_2$ are the reference position and attitude of the hover points and their weights matrices, respectively, and the $\epsilon1_{max}$, $\epsilon2_{max}$ are the extreme value of the two hovering target points' difference of the velocity and the angular velocity to ensure stability. Hence, the ST3M method is proposed to solve the tracking control problem, in which the path consisting of hover points should be found and the target function $J$ should be minimized.

\section{Dynamic Hovering}

In this section, we propose the Smooth Transition Trajectory Tracking Method (ST3M), outlined in Fig. \ref{fig:methodology}. With a RL controller trained for the hardware platform, the VTOL UAV could follow the target path planned on a given trajectory, which can be divided into two stages. The stage I containing dynamic hovering controller training is the main contribution of this paper, while stage II is developed from the state of the art, which are explained in the followings.

The ST3M method is mainly constructed by 2 stages: hover controller training and target path planning. With dynamic modeling, we calculated the relation of thrust force $F_1, F_2, F_3$ (Fig. \ref{fig:dynamic}) with altitude $I$ and position $P$. The objective is to let the RL controller learn to maintain in-flight equilibrium in any given rotor tilt angle $\mu$.

\subsection{Mathematical Modeling and Trim Analysis}
To solve the optimization problem in Equation. \eqref{eq:problem}, deriving the 6-degree-of-freedom (DOf) nonlinear mathematical model of the tri-rotor VTOL UAV and covering the trim analysis in which the values of the control inputs and states are at hover equilibrium. This model aids our reward engineering and physical simulation is needed.
\subsubsection{Nonlinear Equations of Motion}
Define the attitude as the VTOL UAV's Euler angles, including roll $\phi$, pitch $\theta$, and yaw $\psi$. Assuming that the altitude change is limited and, the air density is constant. Therefore, the aerodynamic force and moment produced by the $i$ th rotor represented using the aerodynamic force and moment constants are given by:
\begin{align}
F_i=K_{\mathrm{F}} \Omega_i^2, M_i=K_{\mathrm{M}} \Omega_i^2,
\end{align}
where $K_{\mathrm{F}}$ and $K_{\mathrm{M}}$ are the aerodynamic force and moment constants, respectively. According to Newton’s second law, the components of the aerodynamic force and moment are given by:
\begin{align}
\begin{aligned}
& F_x=K_{\mathrm{F}}(\Omega_2^2 + \Omega_3^2) \sin \mu, F_y=0, \\
& F_z=-K_{\mathrm{F}}\left(\Omega_2^2 \cos \mu+\Omega_3^2 \cos \mu+\Omega_1^2\right), \\
& M_x=-l_3 K_{\mathrm{F}}\left(\Omega_2^2-\Omega_3^2\right) \cos \mu, \\
&M_y=-l_2 K_{\mathrm{F}}\left(\Omega_2^2+\Omega_3^2\right) \cos \mu
+l_1 K_{\mathrm{F}} \Omega_1^2, \\
&M_z=l_3 K_{\mathrm{F}}\left(\Omega_2^2+\Omega_3^2\right) \sin \mu \\
& \quad -K_{\mathrm{M}} \Omega_1^2 +K_{\mathrm{M}}(\Omega_2^2- \Omega_3^2) \cos \mu .
\end{aligned}
\end{align}
\subsubsection{Trim Analysis}
In trim analysis, we select the hover condition as an equilibrium point. by neglecting the gyroscopic moments due to the rotors' inertia, drag forces and moments, the total force $F_{\text {total }}$and $M_{\text {total }}$ that act on the tri-rotor VTOL UAV are given by $\left[F_{\text {total }_x}, F_{\text {total }_y}, F_{\text {total }_z}\right]^{\mathrm{T}}$ and $\left[M_{\text {total }_x}, M_{\text {total }_y}, M_{\text {total }_z}\right]^{\mathrm{T}}$, where:
\begin{align}
\begin{aligned}
F_{\text {total }_x}&=F_x-m g \sin \theta, \\
F_{\text {total }_y}&=F_y + m g \sin \phi \cos \theta, \\
F_{\text {total }_z}&=F_z + m g \cos \phi \cos \theta .
\end{aligned}
\end{align}
Therefore, by ignoring the induced pitching moment by the tilted rotor, the general nonlinear 6-DOf dynamic model can be given by:
\begin{align}
\begin{aligned}
m\left[\begin{array}{c}
\dot{u} \\
\dot{v} \\
\dot{w}
\end{array}\right]&=\left[\begin{array}{c}
F_{\text {total }_x} \\
F_{\text {total }_y} \\
F_{\text {total }_z}
\end{array}\right], \\
\left[\begin{array}{c}
J_x \dot{p} \\
J_y \dot{q} \\
J_z \dot{r}
\end{array}\right]&=\left[\begin{array}{c}
M_{\text {total }_x}-\left(J_z-J_y\right) q r \\
M_{\text {total }_y}-\left(J_x-J_z\right) p r \\
M_{\text {total }_z}-\left(J_y-J_x\right) p q
\end{array}\right], \\
\left[\begin{array}{c}
\dot{\phi} \\
\dot{\theta} \\
\dot{\psi}
\end{array}\right]&=\left[\begin{array}{ccc}
1 & \sin \phi \tan \theta & \cos \phi \tan \theta \\
0 & \cos \phi & -\sin \phi \\
0 & \frac{\sin \phi}{\cos \theta} & \frac{\cos \phi}{\cos \theta}
\end{array}\right]\left[\begin{array}{l}
p \\
q \\
r
\end{array}\right],
\end{aligned}
\label{equ:dynamic}
\end{align}
where $[u, v, w]^{\mathrm{T}}$ and $[p,q,r]^{\mathrm{T}}$ are the velocities and angular body rates in $x$-, $y$-, and $z$-body directions, respectively. In the trim condition, no aerodynamic force or moment is acting on the UAV body because of the absence of forward airspeed. Thus, in the equilibrium point of the transition regime, solving for the unknown inputs and states results in the following analytical expressions of the trim inputs:
\begin{align}
& \phi_{\text {trim }}=0, \theta_{\text {trim }}=\tan ^{-1}\left[\frac{l_2 K_{\mathrm{M}}}{l_3\left(l_1+l_2\right) K_{\mathrm{F}}}\right], \notag \\
& \mu_{\text {trim }}=\tan ^{-1}\left[\frac{l_2 K_{\mathrm{M}}}{l_1 l_3 K_{\mathrm{F}}}\right], \Omega_{1_{\text {trim }}}^2=\frac{l_2 g m}{\left(l_1+l_2\right) K_{\mathrm{F}}} \cos \theta_{\text {trim }}, \notag \\
& \Omega_{2_{\text {trim }}}^2=\frac{l_1 g m}{2\left(l_1+l_2\right) K_{\mathrm{F}}} \frac{\cos \theta_{\text {trim }}}{\cos \mu_{\text {trim }}}, \Omega_{3_{\text {trim }}}=\Omega_{2_{\text {trim }}} .
\label{equ:ideal_parameter}
\end{align}

\subsection{Hover Reward Maximization}

In Stage I of the ST3M algorithm, the HRM algorithm is utilized to develop an RL model capable of precisely controlling the VTOL UAV to hover at the target point within a specified range.

To facilitate observation retrieval, reward calculation, and control command application, a precise digital twin of the VTOL UAV was developed. This model is based on the 6-DOF dynamic model (Equation \eqref{equ:dynamic}) and the physical specifications outlined in Table \ref{tab:specifications}. The ideal hovering state is given by Equation \eqref{equ:ideal_parameter}. The RL controller learns the control strategy by observing the UAV’s attitude $I$, velocity $V$, and angular velocity $\omega$.

The RL controller encourages the UAV to maintain a horizontal attitude and minimize $V$ and $\omega$ for stable hovering, with weights $W_v=0.2$ and $W_\omega=0.1$. Consequently, the output space $O$, reward $R$, and state space $S$ for achieving dynamic hovering are defined as follows:

\begin{align}
\begin{aligned}
    &O=[I, 0.25\sigma(V), \sigma(\omega), \Delta P, S_{prev}],\\
    &R_{step}=proj_{z_e}(z_b)R_{proj} - W_VR_V - W_\omega R_\omega,\\
    &proj_{z_e}(z_b)=((z_b \cdot z_e) / ||z_e||^2) * z_e,\\
    &S=[\Omega_1, \Omega_2, \Omega_3, \mu_1, \mu_2].
\end{aligned}
\end{align}

\begin{algorithm}
 \caption{HRM}
 \begin{algorithmic}[1]
  \REQUIRE Range k; RL Model $f$; UAV Dynamic Model $\lambda$;
  \WHILE{$R \leq R_{max}$}
    \STATE VTOL UAV State $E_\lambda$
    \FOR{$t \in [0, \ldots, T]$}
      \STATE $\{u_t,v_t,w_t,p_t,q_t,r_t\}\Leftarrow E_\lambda$
      \STATE $\Delta P\Leftarrow P_{target}-P$
      \STATE $\omega \Leftarrow [p_t,q_t,r_t]^\mathrm{T}$, $V \Leftarrow [u_t,v_t,w_t]^\mathrm{T}$
      \STATE $O\Leftarrow[I, 0.25\sigma(V), \sigma(\omega), \Delta P, S]$
      \STATE $S\Leftarrow f(O)$
      \STATE $E_\lambda\Leftarrow \lambda(S)$
      \IF{$t\Delta P \leq 1$}
        \STATE $R_{step}\Leftarrow W_{vert}R_{vert} - W_vR_v - W_\omega R_\omega$
        \STATE Update policy, Fit value function maximize $R$
      \ENDIF
    \ENDFOR
    \STATE $P_{target}=P_{target}+Rand(0,k)\textbf{1}_{3\times1}$
  \ENDWHILE
  \RETURN $f$
 \end{algorithmic}
 \label{alg:algorithm1}
\end{algorithm}

The unified control of rotors and servo motors in a single output space demonstrates the system’s coupled control capability. The hover reward maximization (HRM) method, is shown in Algorithm. \ref{alg:algorithm1}, limits the maximum reward and step count with $R_{max}$ and $T$. The tracking target is initially fixed, then randomly moved within a certain distance each time the VTOL UAV completes a cycle. This results in a random walk, with the target point moving to a new position within a $k$ meter range of its current position.

\subsection{Progressive Training and Path Planning}

\begin{figure}[htbp]
\centering
\centerline{\includegraphics[width=\linewidth]{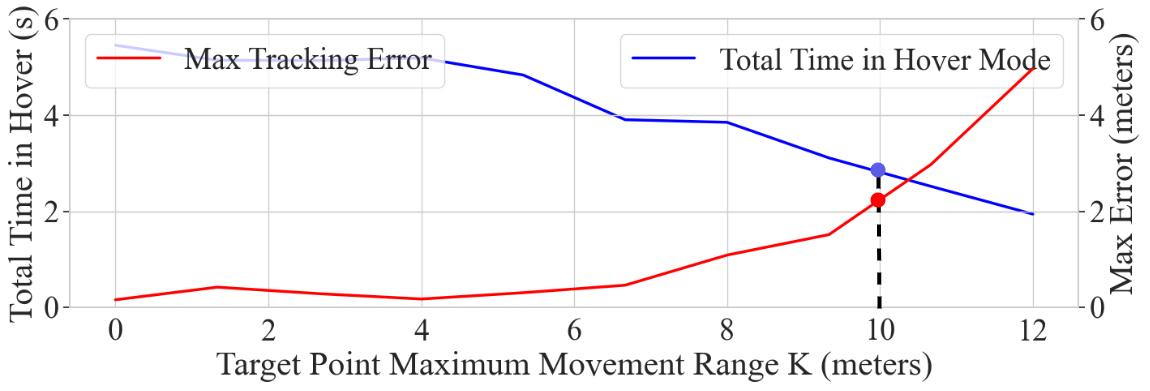}}\caption{The trade-off between target point maximum movement range $K$, tracking time cost and tracking error. The horizontal axis represents the maximum movement range $k$ of the target point. A trade-off between path tracking error and time wasted in hover mode. For less path-tracking error and
less hover time, we choose 10 as the value of $k$, which means the movement of the target point is $Rand(0,10)\textbf{1}_{3\times1}$.}
\label{fig:EfficiencyErrorBalance}
\end{figure}

The training result of the HRM is heavily dependent on the range $k$. Larger $k$ value results in a shorter hover flight time, longer cruise flight time, and faster movement toward the target point for the VTOL UAV. However, it also leads to a rapid increase in maximum tracking error. As depicted in Fig. \ref{fig:EfficiencyErrorBalance}, to maximize overall performance, the optimal maximum movement range $K$ for our model is approximately 10 meters.

\begin{figure}[htbp]
\centering
\centerline{\includegraphics[width=\linewidth]{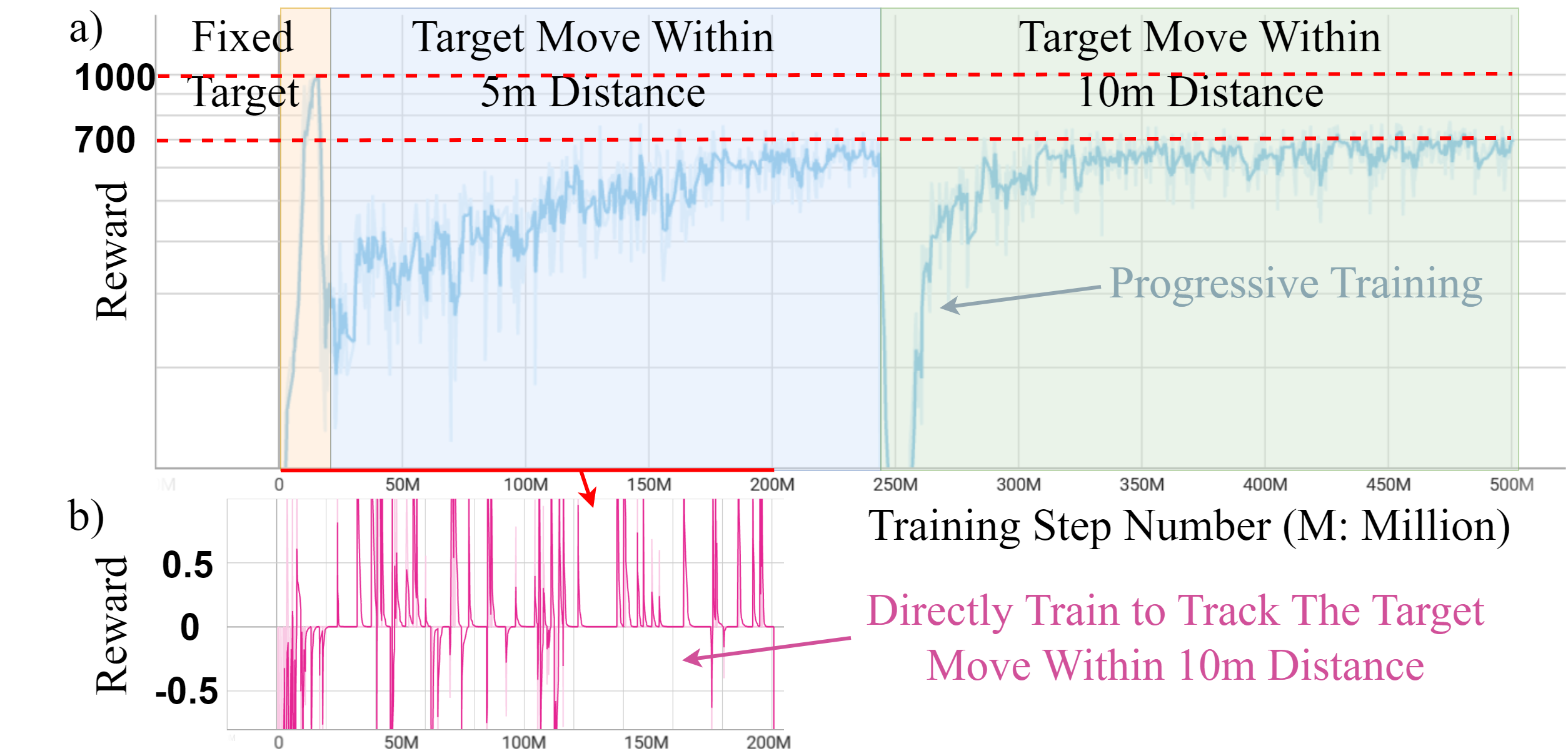}}\caption{Cumulative reward of RL agent progressive training.}
\label{fig:agentreward}
\end{figure}

Fig. \ref{fig:agentreward} illustrates that direct training of the controller to track the target with $k=K=10$ is significantly slower (Pink line in 6b) than hovering on a fixed target (Orange area in 6a), and few effective strategies are learned. This observation confirms the inefficiency of online learning in directly acquiring transition strategies. The training process is divided with a progressive $k=0,5,10$, as shown in Fig. \ref{fig:agentreward}a. This enhances the dynamic tracking ability of the controller.

In stage II, A balanced target path planning algorithm denoted as $BalancePath(T)$ \cite{FU201826} transforms the input trajectory $T$ into a planned path point set $L$ that balances efficiency and accuracy. The RL model then repeatedly executes the try-to-hover maneuver towards this moving point on the planned path, thereby achieving the transition trajectory.

\subsection{Summary}
In summary, we developed a learning-based methodology for VTOL UAV transition control. The HRM method guarantees the dynamic hovering ability. The balanced target path planning method is based on the use of a path-searching algorithm to plan the best path balancing tracking error and time in the cruise mode. The method is effective in two ways: HRM ensures the dynamic hovering ability and progressive training further improves the training time and final performance during transition.

\begin{algorithm}
    \caption{ST3M}
    \begin{algorithmic}[1]
        \STATE \textbf{Input:} Trajectory T; The maximum target movement range K; Dynamic Model $\lambda$;
        \STATE \textbf{// Stage 1: Hover Controller Training}
        \STATE Randomly initialize $f$
        \FOR{$k \in [0, \ldots, K]$}
            \STATE $f \gets \text{HRM}(k, f, \lambda)$
        \ENDFOR
        
        \STATE \textbf{// Stage 2: Balanced Target Path Planning}
        \STATE Planned path $L=\{P_{target}\}^N=BalancePath(T)$

        \FOR{$n \in [0, \ldots, N]$}
            \STATE $\Delta P\Leftarrow L[n]-P$
            \STATE $O\Leftarrow[I, 0.25\sigma(V), \sigma(\omega), \Delta P, S]$
            \STATE $S\Leftarrow f(O)$
            \STATE $E_\lambda\Leftarrow \lambda(S)$
        \ENDFOR
        
    \end{algorithmic}
 \label{alg:algorithm2}
\end{algorithm}

\section{Experiment}
When evaluating RL VTOL UAV controllers' stability, a common approach is to track a target trajectory and assess tracking error, stability, and smoothness. The position and altitude at each simulation step were recorded. For comparison, a dual-loop PID controller with 3 parallel position controllers in the outer loop and 3 parallel PID controllers as reference generators in the inner loop was implemented. The sampling frequency of IMU and Gyroscope in both simulation and real-world model are set to be fixed at 100Hz. For simplicity, the ST3M was implemented and tested using the Proximal Policy Optimization (PPO) algorithm\cite{ppo} as the RL model $f$ and used a predefined target path $L$. These could be replaced with other RL and path-planning algorithms.

\subsection{Simulation Result}
We set the target point maximum movement range $K=10$ and increase by $1$ when the cumulative reward reaches 700. This gives us the following results.

\begin{figure}[htbp]
\centering
\centerline{\includegraphics[width=.9\linewidth]{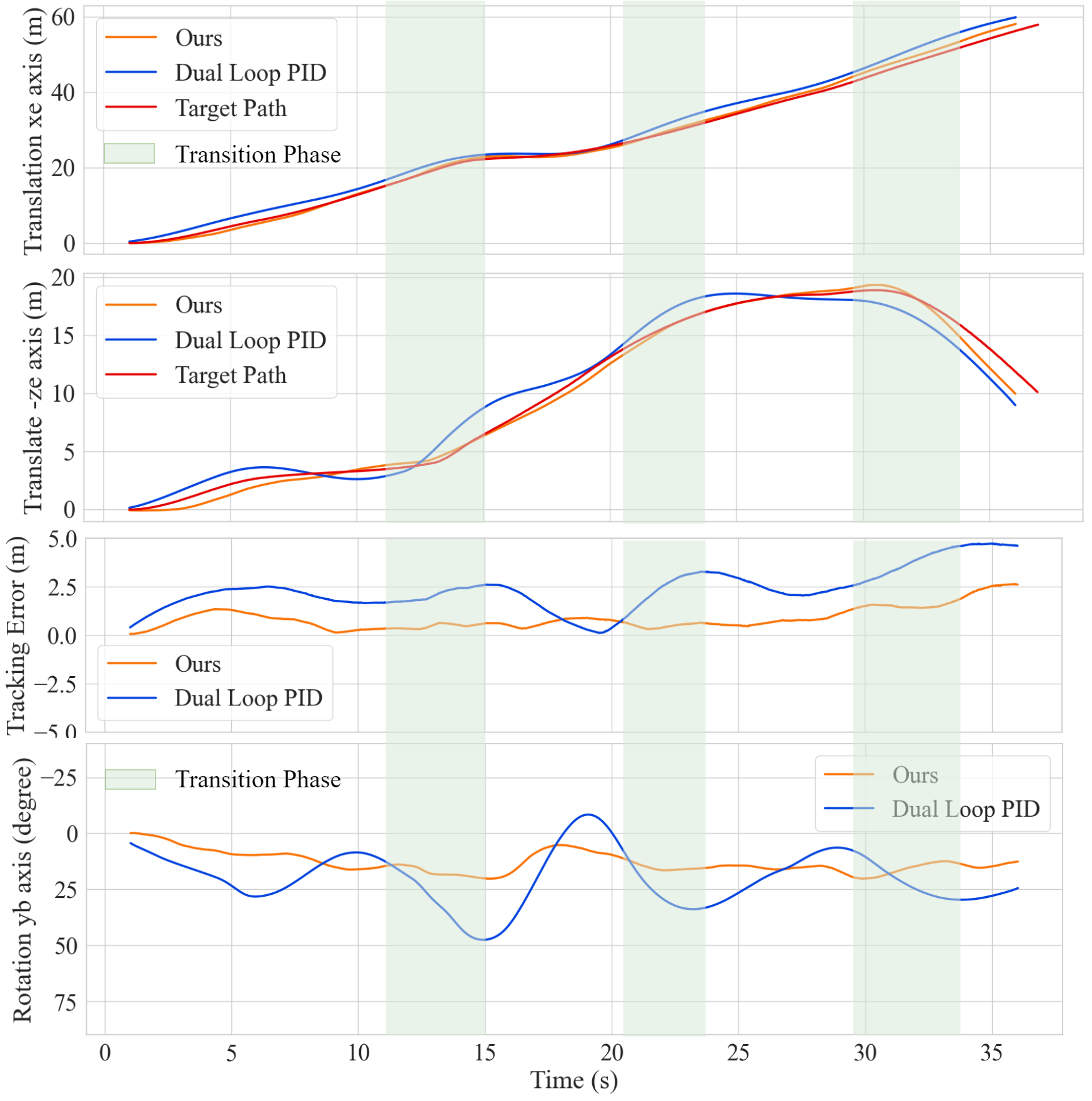}}\caption{VTOL UAV's translation, tracking error and pitch angle (Rotation on axis $y_b$) during the experiment.}
\label{fig:rotate}
\end{figure}

Fig. \ref{fig:rotate} shows the roll and pitch responses of two algorithms. Compared to the Dual Loop PID controller, the RL controller has a smaller shock amplitude, which means the RL controller has a smoother tracking trajectory. It also shows the VTOL UAV's translation on the $x_e$-axis and $-z_e$-axis. We can see that our method fits the target trajectory better with less vibrations. From the quantitative index below we can also conclude that our method outperforms Dual Loop PID with only $42.57\%$ of max pitch angle, $56.03\%$ of max position error, and $36.53\%$ max position error. This shows ST3M's stability and maneuver smoothness.

\begin{table}[h]
    \centering
    \caption{Tilt angle and position errors in simulation}
    \begin{tabular}{cccc}
         \toprule
         Methods & $\overline{\theta}$ (deg) & Max $P_{Err}$ (m) & $\overline{P_{Err}}$ (m)\\
         \midrule
         Dual Loop PID & 47.569 & 4.726 & 2.357\\
         \textbf{ST3M} & \textbf{20.248} & \textbf{2.648} & \textbf{0.861}\\
         \bottomrule
    \end{tabular}
    \label{tab:experiment_results}
\end{table}

\subsection{Real World Verification}
\begin{figure}[htbp]
\centering
\centerline{\includegraphics[width=\linewidth]{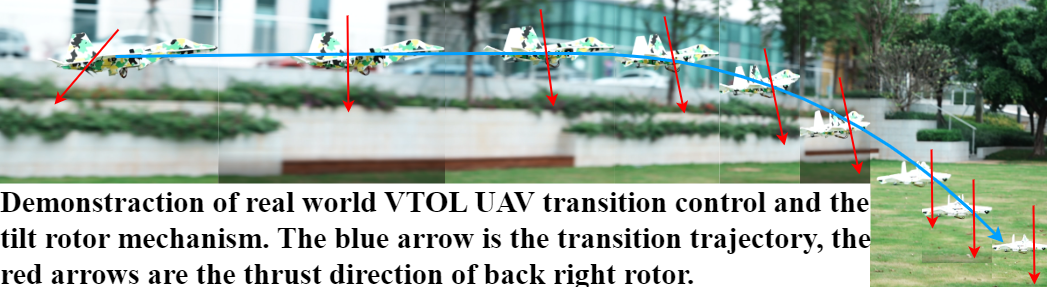}}\caption{Real world verification.}
\label{fig:realworld}
\end{figure}

To verify the usability of the method, a real-world verification platform was built. The real-world VTOL UAV was remotely controlled by a computer. As shown in Fig. \ref{fig:realworld}, a path was pre-planned in which the VTOL UAV transits from cruise flight to hover and finally lands on the ground.

\section{Conclusion}
In conclusion, the ST3M method, utilizing RL, demonstrates remarkable capabilities in stabilizing and maneuvering the smoothness of VTOL UAVs toward target points during transition. This approach enables the RL controller to learn coupled altitude and position control through simulation, by adjusting rotor speed and servo rotor angle directly. Eliminating manual phase switching and dynamic model construction ensures faster adaptation to new hardware and environments, enhances stability and facilitates smoother transitions between hover and cruise phases. The RL controller aligns well with real-world VTOL UAV performance.


\addtolength{\textheight}{-7cm}   










\bibliographystyle{unsrt}
\bibliography{refs}

\end{document}